\documentclass[runningheads]{llncs}
\usepackage[T1]{fontenc}
\usepackage{amsmath,amssymb,amsfonts}
\usepackage{algorithmic}
\usepackage{graphicx}
\usepackage{textcomp}
\usepackage{xcolor}
\usepackage{hhline}
\usepackage{subcaption}
\usepackage{multirow}
\usepackage{colortbl}

\usepackage[utf8x]{inputenc} 

\begin{document}
\title{MoveTouch: Robotic Motion Capturing System with Wearable Tactile Display to Achieve 
\\ Safe HRI}
%
%
\author{Ali Alabbas \inst{1} \and
Miguel Altamirano Cabrera \inst{1} \and
Mohamed Sayed \inst{1} \and \\
Oussama Alyounes \inst{1}, Qian Liu \inst{2}\and
Dzmitry Tsetserukou \inst{1}}
\authorrunning{A. Alabbas et al.}

\institute{Skolkovo Institute of Science and Technology (Skoltech), 121205,  Moscow, Russia
\email{\{ali.alabbas, m.altamirano, mohamed.sayed, oussama.alyounes, d.tsetserukou\}@skoltech.ru}
\\ 
 \and Dalian University of Technology, China, 116024, Dalian City, China\\
\email{qianliu@dlut.edu.cn}}

\titlerunning{MoveTouch }
\maketitle              
\begin{abstract}
The collaborative robot market is flourishing as there is a trend towards simplification, modularity, and increased flexibility on the production line. 
But when humans and robots are collaborating in a shared environment, the safety of humans should be a priority. We introduce a novel wearable robotic system to enhance safety during Human-Robot Interaction (HRI). The proposed wearable robot is designed to hold a fiducial marker and maintain its visibility to a motion capture system, which, in turn, localizes the user’s hand with good accuracy and low latency and provides vibrotactile feedback to the user’s wrist. The vibrotactile feedback guides the user’s hand movement during collaborative tasks in order to increase safety and enhance collaboration efficiency.
A user study was conducted to assess the recognition and discriminability of ten designed vibration patterns applied to the upper (dorsal) and the down (volar) parts of the user’s wrist. The results show that the pattern recognition rate on the volar side was higher, with an average of 75.64\% among all users. Four patterns with a high recognition rate were chosen to be incorporated into our system. A second experiment was carried out to evaluate users' response to the chosen patterns in real-world collaborative tasks. Results show that all participants responded to the patterns correctly, and the average response time for the patterns was between 0.24 and 2.41 seconds.

\end{abstract}
 \begin{figure}[h]
  \centering
  \includegraphics[width=0.7\textwidth]{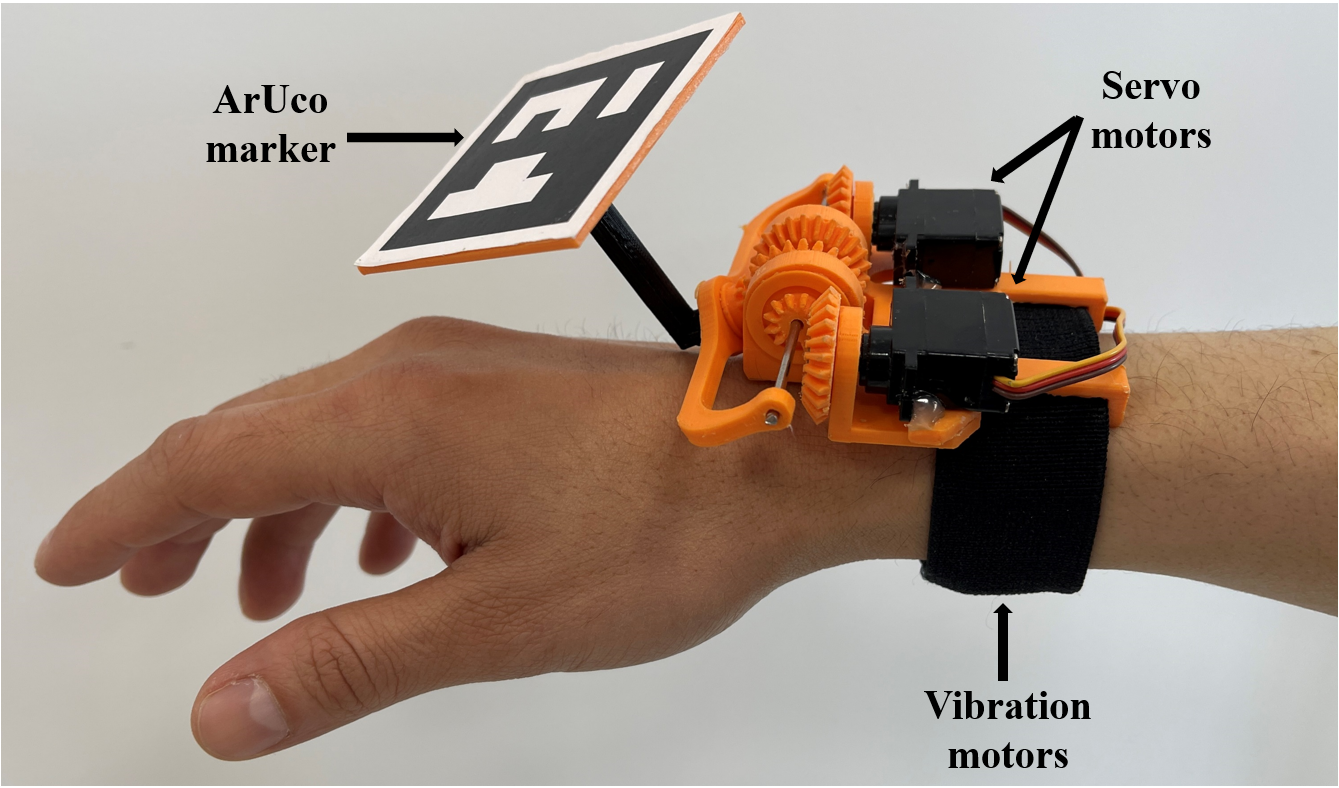}
  \qquad
  \caption{MoveTouch, a novel wearable robot for position tracking and haptic
feedback. An ArUco marker is located at the end effector to adjust its
orientation, while five vibration motors deliver tactile feedback to the user.}
  \label{fig:3d_model}
  \vspace*{-15pt}
\end{figure}

\section{Introduction}
The number of robots in the industry is increasing globally, reaching over 3.9 million robots in factories in 2022, according to the International Federation of Robotics (IFR) \cite{IFR}. Robots are considered helpful associates for carrying out repetitive tasks, while human dexterity can be harnessed in the operation. However, robots are mostly being implemented in the industry as tools, not as companions for humans. To achieve collaboration between humans and robots, the safety of users has to be guaranteed \cite{ZACHARAKI2020104667}.

Haptic feedback is considered one of the most efficient ways to guarantee safety in the application of Human-Robot Interaction (HRI) \cite{seminara2019active}. It works as an excellent notification system for humans \cite{haptic_notification}.

Several studies have focused on enhancing safety in HRI scenarios through tactile devices \cite{pacchierotti2023cutaneous}. Some have included tactile feedback in medical applications \cite{medical}, virtual reality \cite{Haptic_VR} or even in assisting blind people in indoor environments \cite{Haptic_blind}. Tactile feedback can be applied to different parts of the human body, for example, to the fingertips \cite{medical}, to the palm \cite{cabrera2022linkglide} or to the forearm \cite{moriyama2022wearable}, \cite{9551579}, \cite{aevarsson2022vibrotactile}.

In the applications where users need their hands to perform some tasks, providing tactile feedback to the forearm is considered a preferable option since it allows users to use their hands freely. Stanley et al. evaluated ten forms of tactile feedback to the human wrist using five wearable actuators (taper, dragger, squeezer, twister, and vibration) \cite{6226397}. They showed that repeated taps on the subject’s wrist on the side toward which they should turn enhanced the performance of the subjects. However, their work compared between different types of tactile feedback based on the suggested tactile actuators and did not evaluate different patterns of vibration. Chase et al. showed that, by training, it is possible to improve the signal identification in haptic feedback for novices \cite{handbandtraining}.

This research work showed that tactile feedback can enhance the human response and help users perform some actions depending on the haptic feedback that they receive. However, they did not evaluate the different patterns in terms of how effectively users can recognize these patterns. Hong J. et al. developed a wrist haptic device to guide blind people's hands to perform different tasks \cite{hong2017evaluating}. They found, by using 4 and 8 different vibro motors, that single motor feedback was more efficient than interpolated feedback (using more than one motor at a time). However, they did not show the difference between giving feedback to the volar and the dorsal part of the forearm.

ArUcoGlide is a wearable robotic device that we previously developed to ensure that the motion capture system is always able to track the hand of the user \cite{ArUcoGlide}. The main idea of this device is to have a rotatable ArUco marker with two Degrees of Freedom (DOF) to maintain a specific angle with the camera and thus avoid any possible occlusion. In this work, we have enhanced the design of the ArUcoGlide to make it less bulky based on a differential gear train mechanism while ensuring the same performance and adding vibro motors to give tactile feedback to the user's wrist.

This paper aims to study the perception of humans for different tactile patterns exerted on the up (dorsal) and down (volar) parts of the wrist. The tactile feedback is applied through five vibro motors attached to a wearable band. A human study was conducted to compare between the applied patterns depending on the position (dorsal or volar) and the applied frequency. Following the results of the human study, we trained the users to perform some actions depending on the chosen tactile feedback patterns. A human-robot collaborative task was conducted to evaluate the system, including the new design of our motion capture system (ArUcoGlide).

\section{MoveTouch Motion Capture System} \label{System Integration}

The proposed system consists of the following components:

\begin{enumerate}
    \item A wearable robot, dubbed MoveTouch, that is responsible for adjusting the orientation of an ArUco marker held at its end-effector and providing haptic feedback to the user. The movable ArUco marker ensures the visibility of the marker to the motion capture system, avoiding any occlusion due to the movement of the user's hand. The haptic feedback is provided to the user through five vibration motors controlled independently, which are able to generate different vibration patterns. The haptic system acts as a guidance system that tells the user about the optimal movement of their hand during Human-Robot Interaction (HRI) to avoid collisions.

    \item A motion capture (mocap) system consisting of a computer and a single RGB camera that continuously captures a live video stream of the workspace that consists of a collaborative robot UR10 from Universal Robots and the wearable robot.
\end{enumerate}

\subsection{MoveTouch Wearable Robot}

The primary goal of the MoveTouch wearable robot is to assist in determining the user's position and provide vibrotactile guidance, ensuring a secure HRI experience.

\subsubsection{Mechanical Design}

The proposed device is a two-degrees-of-freedom (2-DoFs) wearable robot, based on a differential gear train mechanism, that holds an ArUco marker at its end-effector with the ability to rotate this marker around two perpendicular axes through two servo motors \cite{RussianBook}, \cite{ArticleForMechanism}. Additionally, it includes an array of vibration motors to deliver vibration patterns to the user's wrist. The links, holders, and gears were designed and 3D-printed with PLA material; the 3D CAD model is shown in Fig. \ref{fig:3d_model}. The device maintains the visibility of the ArUco marker to the tracking system by continuously adjusting the motors' angular position to hold the marker in a fixed orientation with respect to the camera coordinate system.

\begin{figure}[htbp]
    \centering
    \begin{subfigure}[b]{0.45\textwidth}
        \centering
        \includegraphics[width=\textwidth]{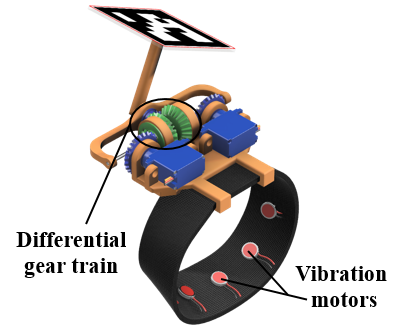}
        \caption{}
        \label{fig:sub1}
    \end{subfigure}
    \hfill
    \begin{subfigure}[b]{0.45\textwidth}
        \centering
        \includegraphics[width=\textwidth]{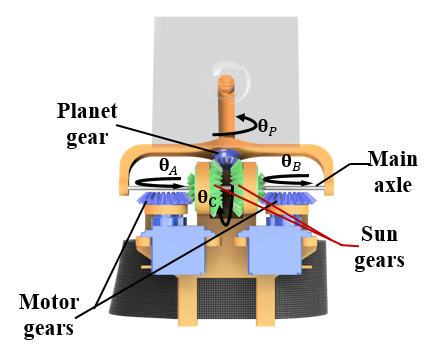}
        \caption{}
        \label{fig:sub2}
    \end{subfigure}
    \caption{MoveTouch design: 3D model perspectives (a) isometric view, (b) top view revealing the gear structure.}
    \label{fig:two_subfigures}
\end{figure}

To achieve the rotation of the end-effector around two perpendicular axes using identical motors positioned side by side, we decided to use a differential gear train mechanism, as shown in Fig. \ref{fig:3d_model}(a). In this figure, the fixed base, the differential gear train, and the rotating Aruco holder are shown.

We denote $\Delta \theta_{A}$ and $\Delta \theta_{B}$ as the rotation angles of the first and second motors, respectively. The rotation angles that the ArUco marker should undergo around the lateral and longitudinal axes are $\Delta \theta_{P}$ and $\Delta \theta_{C}$ which can be seen in Fig. \ref{fig:3d_model}(b). These angles, $\Delta \theta_{P}$ and $\Delta \theta_{C}$, can be acquired from the motion capture system to ensure the ArUco marker maintains a fixed orientation relative to the camera coordinate system \cite{ArUcoGlide}. The equations for determining the required rotation angles of both motors, based on the desired rotation angles for the marker, are as follows:

\begin{equation}
    \Delta \theta_{A} = n_{a} (n_{s}\Delta \theta_{P} + \Delta \theta_{C})
    \label{eq:thetaA}
\end{equation}
\begin{equation}
    \Delta \theta_{B} = n_{b} (n_{s}\Delta \theta_{P} - \Delta \theta_{C}),
    \label{eq:thetaB}
\end{equation}
where $n_{a} = n_{b} = 0.5$ are the gear ratio of the motor gears to the sun gears and $n_{s} = 0.5$ is the gear ratio of the sun gears to the planet gear. The rotation angles $\Delta \theta_{A}$ and $\Delta \theta_{B}$ are acquired via the motion capture system.

We chose to integrate five vibration motors within the system. These motors were chosen due to their small size. The selection of this number takes into consideration two key factors: 1) Enabling a uniform distribution of the vibration motors on one side of the user's wrist (either dorsal or volar), which typically falls within the range of 7.5 -10 cm. 2) This number of vibration motors is sufficient for generating a diverse array of vibrotactile patterns to be employed in user experiments.

\subsubsection{Electronic Design}

The electronic setup includes an ESP32 microcontroller along with two Gotech GS-9025MG servo motors. The system is powered by a Li-Po 7.4 V battery connected through a DC/DC converter to power the microcontroller, the servo driver and the vibration motors. 

Five coin vibration motors, capable of vibrating at frequencies ranging from 10 to 55 Hz, are positioned on the bracelet with a 2 cm separation between them.

The angular position of the servo motors and the activation of the vibration motors are controlled via Bluetooth from a base computer.

\subsection{Motion capture system}

To ensure the safety of users, we need to track the real-time position of the operator within the working space. In this study, the interaction between one user wearing the MoveTouch robot on their wrist and a UR10 robotic manipulator was studied during collaborative task. Given that the MoveTouch marker is situated above the user's hand as shown in Fig. \ref{fig:3d_model}, we suggest that determining the marker's position enables us to locate the hand by applying a transformation from the marker to the center of the hand. In this paper, we will use the terms "user's hand position" and "Movetouch marker" interchangeably.

To achieve a safe interaction, we need to locate the user's hand in the robot coordinate system to make the robot avoid any collision with it. Our proposed mocap system is both cost-efficient and easy to install, utilizing an ArUco marker to track the operator's hand. The system comprises a base computer and an RGB webcam C930e from Logitech mounted on a stand that can be adjusted to capture different angles of the workspace, providing greater flexibility for the user. The basic task of the motion capture system is to transfer the user's hand position into the UR10 robot base coordinate system so that the robot can avoid collision with it. Since the MotionTouch marker, and thereby the user's hand position, can be estimated in the camera coordinate system \cite{garrido2014automatic}, the transformation from the MotionTouch marker to the camera $T_C^{A}$ can be known. By attaching another Aruco marker (we will call it the base marker) at a known transformation from the UR10 robot base, we can also estimate the position of the base of the robot in the camera coordinate system; thus, we can get the transformation $T_C^{B}$ from the UR10 robot base to the camera coordinate system. As a result, we obtain the transformation between the user's hand and the UR10 robot base coordinate system $T_B^A$, and thus the position of the user's hand in the UR10 robot base coordinate system as follows:

\begin{equation}\label{coordinates_transform}
T_B^A = (T_C^{B})^{-1} T_C^{A}
\end{equation}

Calculating the transformation from the base marker to the camera $T_C^{B}$ is required only once before starting the experiments, as the camera will be in the same position throughout the whole experiment. However, if the camera's position or orientation is altered, the process needs to be repeated to derive the correct transformation matrix. Once the transformation matrix $T_B^C$ is determined, we can utilize it to track the position of the MotionTouch marker that is attached to the user's hand, enabling us to locate it within the UR10 robot's coordinate system.

\section{Vibrotactile Guidance System} 

The proposed wearable device, MoveTouch, provides vibrotactile guidance, facilitating human-robot collaboration. This innovative system harnesses tactile sensations to guide users in adjusting their hand position when the collaborative robot approaches and alerts them about potentially risky situations. The system transfers information about how the user should move their hands through vibration patterns exerted on the user's wrist. Ten vibration patterns have been chosen for testing in terms of recognition and discriminability on both the volar and dorsal parts of the user's hand

\subsection{Patterns Design}

Considering that our wearable robot is worn on the wrist of the user, it is logical to incorporate vibrotactile patterns in this specific region. Our initial step is to evaluate the user's ability to recognize vibrotactile patterns applied to their wrist and determine suitable frequencies for these patterns.

\begin{figure}[ht]
  \centering
  \includegraphics[width=\textwidth]{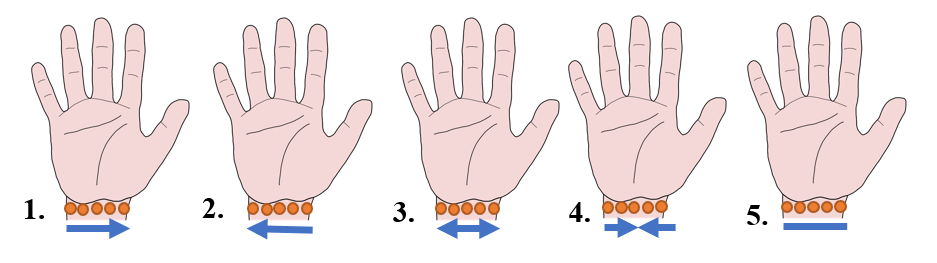}
  \qquad
  \caption{ The designed tactile patterns include: 1) Right-to-left propagation; 2) Left-to-right propagation; 3) Center-to-outside propagation; 4) Outside-to-center propagation; and 5) Simultaneous activation of all vibration motors.}
  \label{patterns}
\end{figure}

Five different patterns were designed using the five vibration motors, each with two different frequencies. The vibration patterns are illustrated in Fig.~\ref{patterns}. In the first pattern, the vibration propagates sequentially through the five vibration patterns from the right side of the user to the left side. The second pattern is similar but the vibration propagation is from left to right. In the third pattern, vibration propagates symmetrically from the center to the outside. Initially, the central vibrator starts vibrating, followed by the simultaneous activation of the two neighboring vibrators. Subsequently, the vibration spreads to the two outermost vibrators. The same methodology for the fourth pattern but the direction is inversed to be from outside to the center. Finally, in the fifth pattern, all vibrators are activated together.  

As for choosing the frequencies of the patterns, we established two levels: high and low. For the high-frequency patterns, each vibration motor was activated for 0.1 seconds, while for the low-frequency patterns, the activation period for each vibrator was 0.2 seconds. Consequently, the high-frequency rates were 2 Hz for the first and second patterns; 3.3 Hz for the third and fourth patterns; and 10 Hz for the fifth pattern. The corresponding low-frequency rates were 1 Hz for the first and second patterns, 1.67 Hz for the third and fourth, and 5 Hz for the fifth pattern. We will label each pattern with H or L at the end to mention the frequency (e.g., 1H means the first pattern with a high frequency). The five patterns with two frequencies provide ten different patterns that we will utilize in our study.

In the following subsection, we present an experimental evaluation to assess the pattern recognition of the vibration patterns by users.

\subsection{Pattern Recognition}

In this experiment, we aimed to assess the ability of the participant to recognize and differentiate between the designed vibrotactile patterns. We invited 11 participants and asked them to tell us which haptic pattern they perceived. Their data was recorded and analyzed.

\subsubsection{Subjects}
Eleven participants, five women and six men, aged from 23 to 34 years (26 $\pm$3.05) took part in the experiment. All selected participants were right-handed. The participants were informed about the experiment and filled out the consent form. 

\subsubsection{User Study Procedure}

Prior to the study, a training session was held to introduce the task, providing users with a detailed explanation of the patterns. Users were asked to wear the MoveTouch robot, and each pattern was rendered three times. A printed piece of paper displaying the patterns was placed in front of the participants throughout the entire training session.

\begin{figure}[htbp]
    \centering
    \begin{subfigure}[b]{0.4\textwidth}
        \centering
        \includegraphics[width=\textwidth]{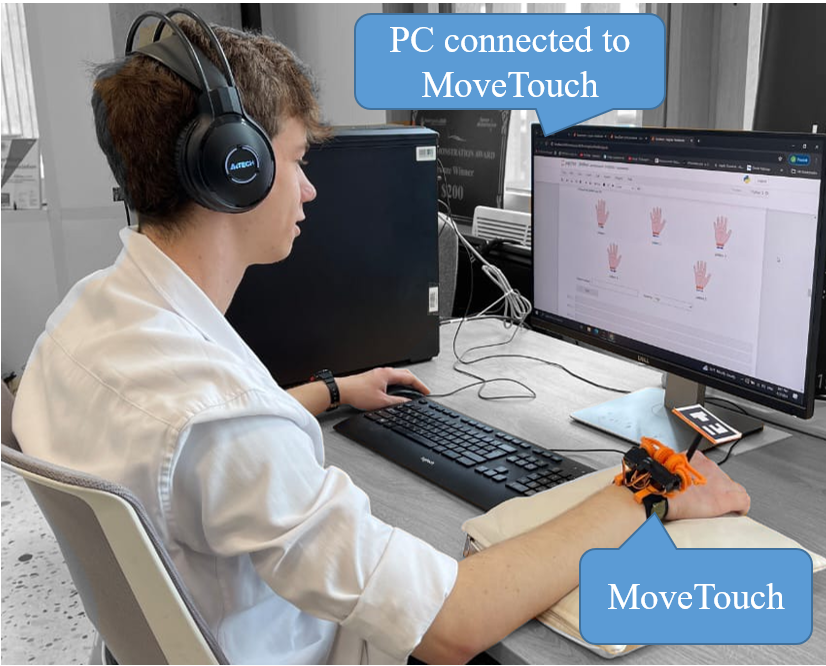}
        \caption{}
        \label{exp1_set}
    \end{subfigure}
    \hfill
    \begin{subfigure}[b]{0.57\textwidth}
        \centering
        \includegraphics[width=\textwidth]{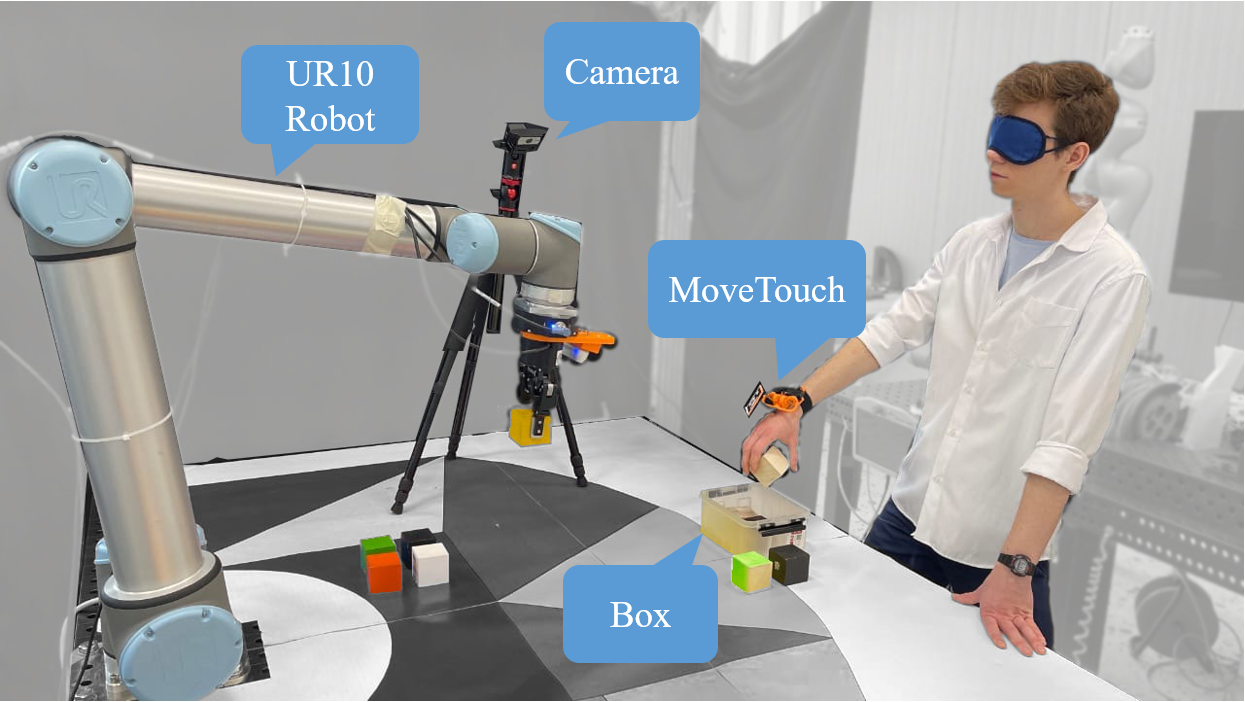}
        \caption{}
        \label{exp2_set}
    \end{subfigure}
    \caption{The experimental setup: a) The pattern recognition experiment setup. b) The system evaluation in a real-world collaborative task setup.}
    \label{experiments}
    \vspace*{-15pt}
\end{figure}

During the study, each user was asked to wear the MoveTouch robot and sit in front of a PC with a graphical user interface (GUI) that allowed them to choose the pattern that they felt. An example of the user study of this experiment is shown in Fig. \ref{exp1_set}.

To identify the most effective part form the wrist to perceive vibration (either dorsal or volar), participants were evaluated twice, wearing the device each time while the vibrators were either on the dorsal or volar part of their wrists. The participants were split into two groups: the first group tried the vibrators on the volar part first, followed by the dorsal part, while the second group followed the opposite sequence. Each of the 10 patterns was replicated five times in random order; thus, 50 patterns were provided to each participant in each evaluation. Participants were asked to wear headphones while playing white noise to prevent them from hearing the vibrations and engaged in the experiment through the GUI interface. They had the option to initiate the vibration pattern when they were ready and select from the displayed patterns the one they felt. During the experiment, participants were not allowed to repeat any pattern.

\subsubsection{Results}
The results for each evaluation (dorsal part and volar part) were analyzed and presented in a confusion matrix. In order to evaluate the statistical significance of the differences between the perceptions of the patterns with different frequencies, we analyzed the results using a single-factor repeated-measures ANOVA, with a chosen significance level of $\alpha<0.05$. The open-source statistical packages Pingouin and Stats models were used for the statistical analysis.

The results of the user perception evaluation by rendering the patterns on the volar part of the wrist are summarized in the confusion matrix (see Table \ref{confusion_down_all}).

\begin{table}[]
\centering{
\caption{Confusion matrix for actual and perceived patterns on the volar (down) part of the wrist}
\label{confusion_down_all}
\setlength{\tabcolsep}{6pt} 
\renewcommand{\arraystretch}{0.95} 
\begin{tabular}{|cc|cccccccccc|}
\hline
\multicolumn{2}{|c|}{\cellcolor[HTML]{FFFFFF}} &
  \multicolumn{10}{c|}{\cellcolor[HTML]{FFFFFF}\textit{Answers   (Predicted Class)}} \\ \hhline{~~----------} 
\multicolumn{2}{|c|}{\multirow{-2}{*}{\cellcolor[HTML]{FFFFFF}\%}} &
  \multicolumn{1}{c|}{\cellcolor[HTML]{FFFFFF}1H} &
  \multicolumn{1}{c|}{\cellcolor[HTML]{FFFFFF}1L} &
  \multicolumn{1}{c|}{\cellcolor[HTML]{FFFFFF}2H} &
  \multicolumn{1}{c|}{\cellcolor[HTML]{FFFFFF}2L} &
  \multicolumn{1}{c|}{\cellcolor[HTML]{FFFFFF}3H} &
  \multicolumn{1}{c|}{3L} &
  \multicolumn{1}{c|}{4H} &
  \multicolumn{1}{c|}{4L} &
  \multicolumn{1}{c|}{5H} &
  \multicolumn{1}{c|}{5L} \\ \hline
\multicolumn{1}{|c|}{\cellcolor[HTML]{FFFFFF}} &
  \cellcolor[HTML]{FFFFFF}1H &
  \multicolumn{1}{l|}{\cellcolor[HTML]{5774AA}\color[HTML]{F9FAFB}0.80} &
  \multicolumn{1}{l|}{\cellcolor[HTML]{E8ECF4}0.11} &
  \multicolumn{1}{l|}{\cellcolor[HTML]{FFFFFF}0.00} &
  \multicolumn{1}{l|}{\cellcolor[HTML]{FFFFFF}0.00} &
  \multicolumn{1}{l|}{\cellcolor[HTML]{F4F6FA}0.05} &
  \multicolumn{1}{l|}{\cellcolor[HTML]{FFFFFF}0.00} &
  \multicolumn{1}{l|}{\cellcolor[HTML]{FCFCFE}0.02} &
  \multicolumn{1}{l|}{\cellcolor[HTML]{FCFCFE}0.02} &
  \multicolumn{1}{l|}{\cellcolor[HTML]{FFFFFF}0.00} &
  \cellcolor[HTML]{FFFFFF}0.00 \\ \hhline{~-----------}
\multicolumn{1}{|c|}{\cellcolor[HTML]{FFFFFF}} &
  \cellcolor[HTML]{FFFFFF}1L &
  \multicolumn{1}{l|}{\cellcolor[HTML]{F8F9FC}0.04} &
  \multicolumn{1}{l|}{\cellcolor[HTML]{345898}\color[HTML]{F9FAFB}0.96} &
  \multicolumn{1}{l|}{\cellcolor[HTML]{FFFFFF}0.00} &
  \multicolumn{1}{l|}{\cellcolor[HTML]{FFFFFF}0.00} &
  \multicolumn{1}{l|}{\cellcolor[HTML]{FFFFFF}0.00} &
  \multicolumn{1}{l|}{\cellcolor[HTML]{FFFFFF}0.00} &
  \multicolumn{1}{l|}{\cellcolor[HTML]{FFFFFF}0.00} &
  \multicolumn{1}{l|}{\cellcolor[HTML]{FFFFFF}0.00} &
  \multicolumn{1}{l|}{\cellcolor[HTML]{FFFFFF}0.00} &
  \cellcolor[HTML]{FFFFFF}0.00 \\ \hhline{~-----------}
\multicolumn{1}{|c|}{\cellcolor[HTML]{FFFFFF}} &
  \cellcolor[HTML]{FFFFFF}2H &
  \multicolumn{1}{l|}{\cellcolor[HTML]{FCFCFE}0.02} &
  \multicolumn{1}{l|}{\cellcolor[HTML]{FFFFFF}0.00} &
  \multicolumn{1}{l|}{\cellcolor[HTML]{6E87B6}\color[HTML]{F9FAFB}0.69} &
  \multicolumn{1}{l|}{\cellcolor[HTML]{E1E6F0}0.15} &
  \multicolumn{1}{l|}{\cellcolor[HTML]{F4F6FA}0.05} &
  \multicolumn{1}{l|}{\cellcolor[HTML]{FCFCFE}0.02} &
  \multicolumn{1}{l|}{\cellcolor[HTML]{F8F9FC}0.04} &
  \multicolumn{1}{l|}{\cellcolor[HTML]{FCFCFE}0.02} &
  \multicolumn{1}{l|}{\cellcolor[HTML]{FFFFFF}0.00} &
  \cellcolor[HTML]{FCFCFE}0.02 \\ \hhline{~-----------}
\multicolumn{1}{|c|}{\cellcolor[HTML]{FFFFFF}} &
  \cellcolor[HTML]{FFFFFF}2L &
  \multicolumn{1}{l|}{\cellcolor[HTML]{FFFFFF}0.00} &
  \multicolumn{1}{l|}{\cellcolor[HTML]{FFFFFF}0.00} &
  \multicolumn{1}{l|}{\cellcolor[HTML]{FCFCFE}0.02} &
  \multicolumn{1}{l|}{\cellcolor[HTML]{305496}\color[HTML]{F9FAFB}0.98} &
  \multicolumn{1}{l|}{\cellcolor[HTML]{FFFFFF}0.00} &
  \multicolumn{1}{l|}{\cellcolor[HTML]{FFFFFF}0.00} &
  \multicolumn{1}{l|}{\cellcolor[HTML]{FFFFFF}0.00} &
  \multicolumn{1}{l|}{\cellcolor[HTML]{FFFFFF}0.00} &
  \multicolumn{1}{l|}{\cellcolor[HTML]{FFFFFF}0.00} &
  \cellcolor[HTML]{FFFFFF}0.00 \\ \hhline{~-----------}
\multicolumn{1}{|c|}{\cellcolor[HTML]{FFFFFF}} &
  \cellcolor[HTML]{FFFFFF}3H &
  \multicolumn{1}{l|}{\cellcolor[HTML]{FCFCFE}0.02} &
  \multicolumn{1}{l|}{\cellcolor[HTML]{FFFFFF}0.00} &
  \multicolumn{1}{l|}{\cellcolor[HTML]{FFFFFF}0.00} &
  \multicolumn{1}{l|}{\cellcolor[HTML]{FFFFFF}0.00} &
  \multicolumn{1}{l|}{\cellcolor[HTML]{AFBDD7}0.38} &
  \multicolumn{1}{l|}{\cellcolor[HTML]{F4F6FA}0.05} &
  \multicolumn{1}{l|}{\cellcolor[HTML]{A7B7D3}0.42} &
  \multicolumn{1}{l|}{\cellcolor[HTML]{F0F3F8}0.07} &
  \multicolumn{1}{l|}{\cellcolor[HTML]{F4F6FA}0.05} &
  \cellcolor[HTML]{FFFFFF}0.00 \\ \hhline{~-----------} 
\multicolumn{1}{|c|}{\cellcolor[HTML]{FFFFFF}} &
  3L &
  \multicolumn{1}{l|}{\cellcolor[HTML]{FFFFFF}0.00} &
  \multicolumn{1}{l|}{\cellcolor[HTML]{F8F9FC}0.04} &
  \multicolumn{1}{l|}{\cellcolor[HTML]{FFFFFF}0.00} &
  \multicolumn{1}{l|}{\cellcolor[HTML]{FFFFFF}0.00} &
  \multicolumn{1}{l|}{\cellcolor[HTML]{F4F6FA}0.05} &
  \multicolumn{1}{l|}{\cellcolor[HTML]{5371A8}\color[HTML]{F9FAFB}0.82} &
  \multicolumn{1}{l|}{\cellcolor[HTML]{FFFFFF}0.00} &
  \multicolumn{1}{l|}{\cellcolor[HTML]{ECF0F6}0.09} &
  \multicolumn{1}{l|}{\cellcolor[HTML]{FFFFFF}0.00} &
  \cellcolor[HTML]{FFFFFF}0.00 \\ \hhline{~-----------}
\multicolumn{1}{|c|}{\cellcolor[HTML]{FFFFFF}} &
  4H &
  \multicolumn{1}{l|}{\cellcolor[HTML]{F0F3F8}0.07} &
  \multicolumn{1}{l|}{\cellcolor[HTML]{FFFFFF}0.00} &
  \multicolumn{1}{l|}{\cellcolor[HTML]{FFFFFF}0.00} &
  \multicolumn{1}{l|}{\cellcolor[HTML]{FFFFFF}0.00} &
  \multicolumn{1}{l|}{\cellcolor[HTML]{E5E9F2}0.13} &
  \multicolumn{1}{l|}{\cellcolor[HTML]{FCFCFE}0.02} &
  \multicolumn{1}{l|}{\cellcolor[HTML]{7991BB}\color[HTML]{F9FAFB}0.64} &
  \multicolumn{1}{l|}{\cellcolor[HTML]{F0F3F8}0.07} &
  \multicolumn{1}{l|}{\cellcolor[HTML]{F0F3F8}0.07} &
  \cellcolor[HTML]{FFFFFF}0.00 \\ \hhline{~-----------}
\multicolumn{1}{|c|}{\cellcolor[HTML]{FFFFFF}} &
  4L &
  \multicolumn{1}{l|}{\cellcolor[HTML]{FFFFFF}0.00} &
  \multicolumn{1}{l|}{\cellcolor[HTML]{FFFFFF}0.00} &
  \multicolumn{1}{l|}{\cellcolor[HTML]{FFFFFF}0.00} &
  \multicolumn{1}{l|}{\cellcolor[HTML]{FCFCFE}0.02} &
  \multicolumn{1}{l|}{\cellcolor[HTML]{FFFFFF}0.00} &
  \multicolumn{1}{l|}{\cellcolor[HTML]{F0F3F8}0.07} &
  \multicolumn{1}{l|}{\cellcolor[HTML]{E5E9F2}0.13} &
  \multicolumn{1}{l|}{\cellcolor[HTML]{5B77AC}\color[HTML]{F9FAFB}0.78} &
  \multicolumn{1}{l|}{\cellcolor[HTML]{FFFFFF}0.00} &
  \cellcolor[HTML]{FFFFFF}0.00 \\ \hhline{~-----------}
\multicolumn{1}{|c|}{\cellcolor[HTML]{FFFFFF}} &
  5H &
  \multicolumn{1}{l|}{\cellcolor[HTML]{FFFFFF}0.00} &
  \multicolumn{1}{l|}{\cellcolor[HTML]{FFFFFF}0.00} &
  \multicolumn{1}{l|}{\cellcolor[HTML]{FFFFFF}0.00} &
  \multicolumn{1}{l|}{\cellcolor[HTML]{FFFFFF}0.00} &
  \multicolumn{1}{l|}{\cellcolor[HTML]{E8ECF4}0.11} &
  \multicolumn{1}{l|}{\cellcolor[HTML]{FFFFFF}0.00} &
  \multicolumn{1}{l|}{\cellcolor[HTML]{FCFCFE}0.02} &
  \multicolumn{1}{l|}{\cellcolor[HTML]{FFFFFF}0.00} &
  \multicolumn{1}{l|}{\cellcolor[HTML]{5371A8}\color[HTML]{F9FAFB}0.82} &
  \cellcolor[HTML]{F4F6FA}0.05 \\ \hhline{~-----------}
\multicolumn{1}{|c|}{\multirow{-10}{*}{\cellcolor[HTML]{FFFFFF}\textit{\rotatebox{90}{Patterns}}}} &
  5L &
  \multicolumn{1}{l|}{\cellcolor[HTML]{FCFCFE}0.02} &
  \multicolumn{1}{l|}{\cellcolor[HTML]{FFFFFF}0.00} &
  \multicolumn{1}{l|}{\cellcolor[HTML]{FFFFFF}0.00} &
  \multicolumn{1}{l|}{\cellcolor[HTML]{FFFFFF}0.00} &
  \multicolumn{1}{l|}{\cellcolor[HTML]{F8F9FC}0.04} &
  \multicolumn{1}{l|}{\cellcolor[HTML]{FCFCFE}0.02} &
  \multicolumn{1}{l|}{\cellcolor[HTML]{F8F9FC}0.04} &
  \multicolumn{1}{l|}{\cellcolor[HTML]{FCFCFE}0.02} &
  \multicolumn{1}{l|}{\cellcolor[HTML]{D9E0EC}0.18} &
  \cellcolor[HTML]{6E87B6}\color[HTML]{F9FAFB}0.69 \\ \hline

\end{tabular}}
\end{table}

According to the ANOVA results, there is a statistically significant difference in the recognition rates for the different patterns on the volar part of the wrist: $F(9,100) = 5.78, p = 2\cdot10^{-6}$. The ANOVA showed that the patterns and frequencies significantly influenced the perceptions of the users.

The paired t-tests with one-step Bonferroni correction showed statistically significant differences between the patterns 1H and 3H ($p=4.95\cdot10^{-3} < 0.05$), 1L and 3H ($p = 1\cdot10^{-7} < 0.05$), 3H and 4L ($p = 7.3\cdot10^{-5} < 0.05$), 3H and 5H ($p = 3.68\cdot10^{-2} < 0.05$), 2L and 3H ($p =1\cdot10^{-7} < 0.05$), 2L and 4L ($p = 1.40\cdot10^{-2} < 0.05$), and 3H and 3L ($p = 2.24\cdot10^{-3} < 0.05$).

The results of the human perception evaluation by rendering the patterns on the dorsal part of the wrist are summarized in the second confusion matrix (see Table \ref{confusion_up_all}).


\begin{table}[]
\centering{
\caption{Confusion matrix for actual and perceived patterns on the dorsal (up) part of the wrist}
\label{confusion_up_all}
\setlength{\tabcolsep}{6pt} 
\renewcommand{\arraystretch}{0.95} 
\begin{tabular}{|cc|cccccccccc|}
\hline
 
\multicolumn{2}{|c|}{} &
  \multicolumn{10}{c|}{\textit{Answers   (Predicted Class)}} \\ \cline{3-12} 
\multicolumn{2}{|c|}{\multirow{-2}{*}{\%}} &
  \multicolumn{1}{c|}{1H} &
  \multicolumn{1}{c|}{1L} &
  \multicolumn{1}{c|}{2H} &
  \multicolumn{1}{c|}{2L} &
  \multicolumn{1}{c|}{3H} &
  \multicolumn{1}{c|}{3L} &
  \multicolumn{1}{c|}{4H} &
  \multicolumn{1}{c|}{4L} &
  \multicolumn{1}{c|}{5H} &
  \multicolumn{1}{c|}{5L} \\ \hline
 
\multicolumn{1}{|c|}{} &
  1H &
  \multicolumn{1}{l|}{\cellcolor[HTML]{40629F}\color[HTML]{F9FAFB}0.87} &
  \multicolumn{1}{l|}{\cellcolor[HTML]{F0F2F7}0.07} &
  \multicolumn{1}{l|}{0.00} &
  \multicolumn{1}{l|}{0.00} &
  \multicolumn{1}{l|}{0.00} &
  \multicolumn{1}{l|}{0.02} &
  \multicolumn{1}{l|}{0.02} &
  \multicolumn{1}{l|}{0.02} &
  \multicolumn{1}{l|}{0.00} &
  0.00 \\ \hhline{~-----------} 
 
\multicolumn{1}{|c|}{} &
  1L &
  \multicolumn{1}{l|}{\cellcolor[HTML]{F4F6F9}0.05} &
  \multicolumn{1}{l|}{\cellcolor[HTML]{305496}\color[HTML]{F9FAFB}0.95} &
  \multicolumn{1}{l|}{0.00} &
  \multicolumn{1}{l|}{0.00} &
  \multicolumn{1}{l|}{0.00} &
  \multicolumn{1}{l|}{0.00} &
  \multicolumn{1}{l|}{0.00} &
  \multicolumn{1}{l|}{0.00} &
  \multicolumn{1}{l|}{0.00} &
  0.00 \\ \hhline{~-----------} 
\multicolumn{1}{|c|}{} &
  2H &
  \multicolumn{1}{l|}{0.00} &
  \multicolumn{1}{l|}{0.02} &
  \multicolumn{1}{l|}{\cellcolor[HTML]{5875AB}\color[HTML]{F9FAFB}0.76} &
  \multicolumn{1}{l|}{\cellcolor[HTML]{F0F2F7}0.07} &
  \multicolumn{1}{l|}{\cellcolor[HTML]{ECEFF5}0.09} &
  \multicolumn{1}{l|}{0.02} &
  \multicolumn{1}{l|}{0.02} &
  \multicolumn{1}{l|}{0.02} &
  \multicolumn{1}{l|}{0.00} &
  0.00 \\  \hhline{~-----------}

\multicolumn{1}{|c|}{} &
  2L &
  \multicolumn{1}{l|}{0.00} &
  \multicolumn{1}{l|}{0.00} &
  \multicolumn{1}{l|}{\cellcolor[HTML]{F4F6F9}0.05} &
  \multicolumn{1}{l|}{\cellcolor[HTML]{305496}\color[HTML]{F9FAFB}0.95} &
  \multicolumn{1}{l|}{0.00} &
  \multicolumn{1}{l|}{0.00} &
  \multicolumn{1}{l|}{0.00} &
  \multicolumn{1}{l|}{0.00} &
  \multicolumn{1}{l|}{0.00} &
  0.00 \\ \hhline{~-----------}

\multicolumn{1}{|c|}{} &
  3H &
  \multicolumn{1}{l|}{0.00} &
  \multicolumn{1}{l|}{0.00} &
  \multicolumn{1}{l|}{0.00} &
  \multicolumn{1}{l|}{0.00} &
  \multicolumn{1}{l|}{\cellcolor[HTML]{9CADCD}0.45} &
  \multicolumn{1}{l|}{\cellcolor[HTML]{F8F9FB}0.04} &
  \multicolumn{1}{l|}{\cellcolor[HTML]{ACBAD5}0.38} &
  \multicolumn{1}{l|}{\cellcolor[HTML]{F8F9FB}0.04} &
  \multicolumn{1}{l|}{\cellcolor[HTML]{F4F6F9}0.05} &
  \cellcolor[HTML]{F8F9FB}0.04 \\ \hhline{~-----------}
\multicolumn{1}{|c|}{} &
  3L &
  \multicolumn{1}{l|}{0.02} &
  \multicolumn{1}{l|}{0.00} &
  \multicolumn{1}{l|}{0.00} &
  \multicolumn{1}{l|}{\cellcolor[HTML]{F4F6F9}0.05} &
  \multicolumn{1}{l|}{\cellcolor[HTML]{F8F9FB}0.04} &
  \multicolumn{1}{l|}{\cellcolor[HTML]{647FB1}\color[HTML]{F9FAFB}0.71} &
  \multicolumn{1}{l|}{0.00} &
  \multicolumn{1}{l|}{\cellcolor[HTML]{D8DFEB}0.18} &
  \multicolumn{1}{l|}{0.00} &
  0.00 \\ \hhline{~-----------}
\multicolumn{1}{|c|}{} &
  4H &
  \multicolumn{1}{l|}{\cellcolor[HTML]{F4F6F9}0.05} &
  \multicolumn{1}{l|}{0.00} &
  \multicolumn{1}{l|}{0.02} &
  \multicolumn{1}{l|}{0.00} &
  \multicolumn{1}{l|}{\cellcolor[HTML]{B4C1D9}0.35} &
  \multicolumn{1}{l|}{0.02} &
  \multicolumn{1}{l|}{\cellcolor[HTML]{A4B4D1}0.42} &
  \multicolumn{1}{l|}{0.02} &
  \multicolumn{1}{l|}{\cellcolor[HTML]{E4E8F1}0.13} &
  0.00 \\ \hhline{~-----------} 
\multicolumn{1}{|c|}{} &
  4L &
  \multicolumn{1}{l|}{\cellcolor[HTML]{F4F6F9}0.05} &
  \multicolumn{1}{l|}{0.02} &
  \multicolumn{1}{l|}{0.00} &
  \multicolumn{1}{l|}{0.02} &
  \multicolumn{1}{l|}{\cellcolor[HTML]{F4F6F9}0.05} &
  \multicolumn{1}{l|}{\cellcolor[HTML]{D8DFEB}0.18} &
  \multicolumn{1}{l|}{\cellcolor[HTML]{E4E8F1}0.13} &
  \multicolumn{1}{l|}{\cellcolor[HTML]{94A7C9}0.49} &
  \multicolumn{1}{l|}{0.00} &
  \cellcolor[HTML]{F4F6F9}0.05 \\ \hhline{~-----------} 
\multicolumn{1}{|c|}{} &
  5H &
  \multicolumn{1}{l|}{0.00} &
  \multicolumn{1}{l|}{0.00} &
  \multicolumn{1}{l|}{0.00} &
  \multicolumn{1}{l|}{0.00} &
  \multicolumn{1}{l|}{\cellcolor[HTML]{F8F9FB}0.04} &
  \multicolumn{1}{l|}{0.00} &
  \multicolumn{1}{l|}{\cellcolor[HTML]{F4F6F9}0.05} &
  \multicolumn{1}{l|}{0.00} &
  \multicolumn{1}{l|}{\cellcolor[HTML]{5472A9}\color[HTML]{F9FAFB}0.78} &
  \cellcolor[HTML]{E4E8F1}0.13 \\ \hhline{~-----------} 
\multicolumn{1}{|c|}{\multirow{-10}{*}{\textit{\rotatebox{90}{Patterns}}}} &
  5L &
  \multicolumn{1}{l|}{0.00} &
  \multicolumn{1}{l|}{0.00} &
  \multicolumn{1}{l|}{0.00} &
  \multicolumn{1}{l|}{0.00} &
  \multicolumn{1}{l|}{\cellcolor[HTML]{F8F9FB}0.04} &
  \multicolumn{1}{l|}{0.00} &
  \multicolumn{1}{l|}{\cellcolor[HTML]{F8F9FB}0.04} &
  \multicolumn{1}{l|}{0.02} &
  \multicolumn{1}{l|}{\cellcolor[HTML]{D4DBE9}0.20} &
  \cellcolor[HTML]{647FB1}\color[HTML]{F9FAFB}0.71 \\ \hline
\end{tabular}}
\end{table}

According to the ANOVA results, there is a statistically significant difference in the recognition rates for the different patterns on the dorsal part of the wrist: $F(9,100) =  7.5724, p =2.158\cdot10^{-8}$. The ANOVA showed that the patterns and frequencies significantly influenced the perceptions of the users.

The paired t-tests with one-step Bonferroni correction showed statistically significant differences between the patterns 1H and 3H ($p=0.0026 < 0.05$), 1H and 4H ($p = 0.0176< 0.05$), 1H and 4L ($p = 0.001 < 0.05$), 1L and 3H ($p = 0.0003 < 0.05$), 1L and 4H ($p = 0.003 < 0.05$), 1L and 4L ($p = 8\cdot10^{-5} < 0.05$), 2L and 3H ($p = 0.0001 < 0.05$), and 2L and 4H ($p = 0.002 < 0.05$).

From the two confusion matrices, the average recognition rate of all patterns on the dorsal part of the wrist was 66.6\%, while on the volar side of the wrist, it was 75.64\%.

\subsubsection{Conclusion}

We can notice that almost all the patterns are better recognized on the lower part of the wrist; thus, this position of the vibration motors was selected. These patterns include 1H, 1L, 2L, 3L, 4L, and 5H. However, to avoid user confusion, similar patterns like 1H and 1L are not included together. Instead, the pattern with the highest recognition, which is 1L, is selected. Consequently, patterns 1L, 2L, 3L, and 5H will used for further studies.

\subsection{System Evaluation in a Collaborative Task} 

In this evaluation, the system was integrated into a real collaborative task to assess its effectiveness in enhancing the overall user safety and guiding users in a desired direction. The four selected patterns were integrated to direct users on hand movements as follows: Pattern 1L indicates moving the hand to the right, pattern 2L indicates moving the hand to the left, pattern 3L indicates moving the hand down, and pattern 5H indicates moving the hand back.

\subsubsection{Subjects}
Eight participants, comprising 5 men and 3 women, were randomly selected from the set of participants who took part in the first experiment.

\subsubsection{Experimental setup}

The participants were asked to stand in front of a Siegmund welding table, where a UR10 robot was performing a pick-and-place task of cubes positioned at specific locations on the table. The participants wore the MoveTouch on their right wrist and utilized an eye cover to eliminate visual feedback about the location of the robot's Tool Center Point (TCP). Participants were also performing a pick-and-place task in the same box that the robot was using. A surveillance camera was placed on a tripod near the table, where it could monitor the entire workspace. Additionally, An observer was present during the experiment, poised to activate the emergency button to halt the robot in case of any unforeseen issues. The experimental setup is shown in Fig. \ref{experiments}(b).

\subsubsection{Collaborative Task Procedure}
Before the experiment starts, each participant underwent a training session covering the four patterns and the corresponding movements associated with each pattern.

The robot picked cubes from certain positions on the table and placed them in a box near the participant. The participants were asked initially to touch and familiarize themselves with the box to memorize its position. Throughout the experiment, participants were asked to pick cubes near the box and place them inside the box.
The whole experiment was conducted while the user's eyes were covered. 

During collaboration, when the robot's TCP is moving towards the participant's hand and gets to a distance closer than 40 cm (the activation area), the system activates a certain haptic pattern on the participant's wrist to make participants move their hand in response to each pattern as follows: 1) If pattern 1L is rendered, the participant should move their hand to the left. 2) When the pattern 2L is rendered, the participant should move the hand to the right. 3) For the pattern 3L, the movement should be downward. 4) For pattern 5H, the participant should move their hand backward.

For safety concerns, the robot was programmed to halt when its TCP came within a critical threshold (25 cm) from the participant's hand and continue when the participant's hand moved outside the critical area.

\begin{figure}[ht]
  \centering
  \includegraphics[width=0.8\textwidth]{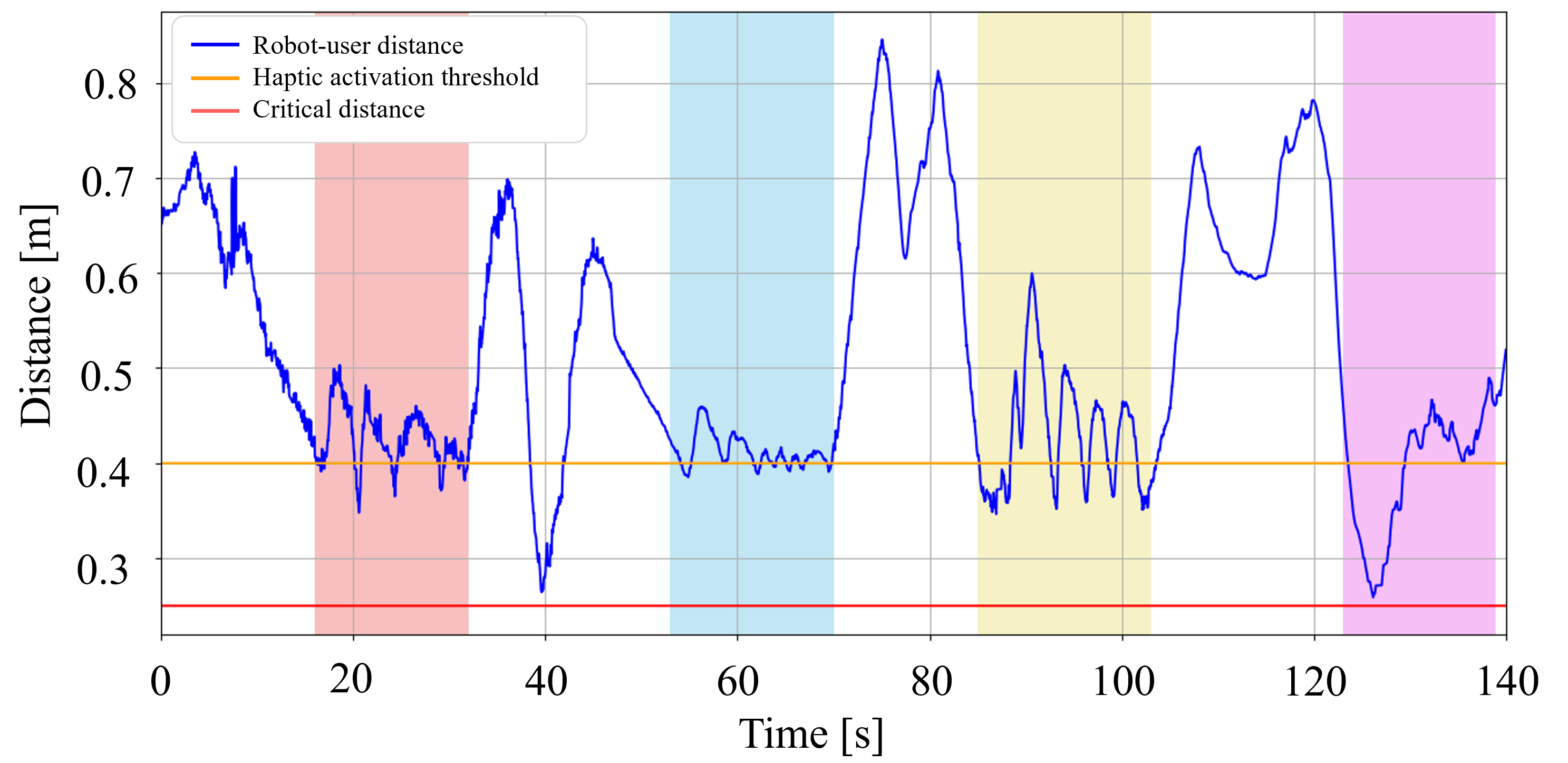}
  \qquad
  \caption{ The distance between the participant and the robot's TCP during one experiment. The red line is the critical distance, while the orange line represents the haptic activation distance. The four highlighted areas are the areas where the haptic patterns were activated.}
  \label{dist}
  \vspace*{-20pt}
\end{figure}

\subsubsection{Results}

The distance between the participant's hand and the robot's TCP is illustrated in Fig.~\ref{dist}. We can observe that the critical distance was preserved throughout the experiment, which indicates safer collaboration, taking into account that the user was doing the task while eye-covered. All participants successfully responded to the haptic patterns,  except for one instance where the participant responded to the pattern 5H by moving his hand downward instead of backward. We can see the effective response to the haptic patterns in the highlighted areas in Fig.~\ref{dist}. These highlighted areas correspond to moments when the robot approached the participant. For example, in the blue highlighted area, the participant attempted to continue the task and return his hand near the box many times, but the robot was still approaching, triggering the haptic pattern multiple times. Consequently, the participant moves their hand again, resulting in an oscillating distance between the robot and the user. 

\begin{figure}[ht]
  \centering
  \includegraphics[width=0.7\textwidth]{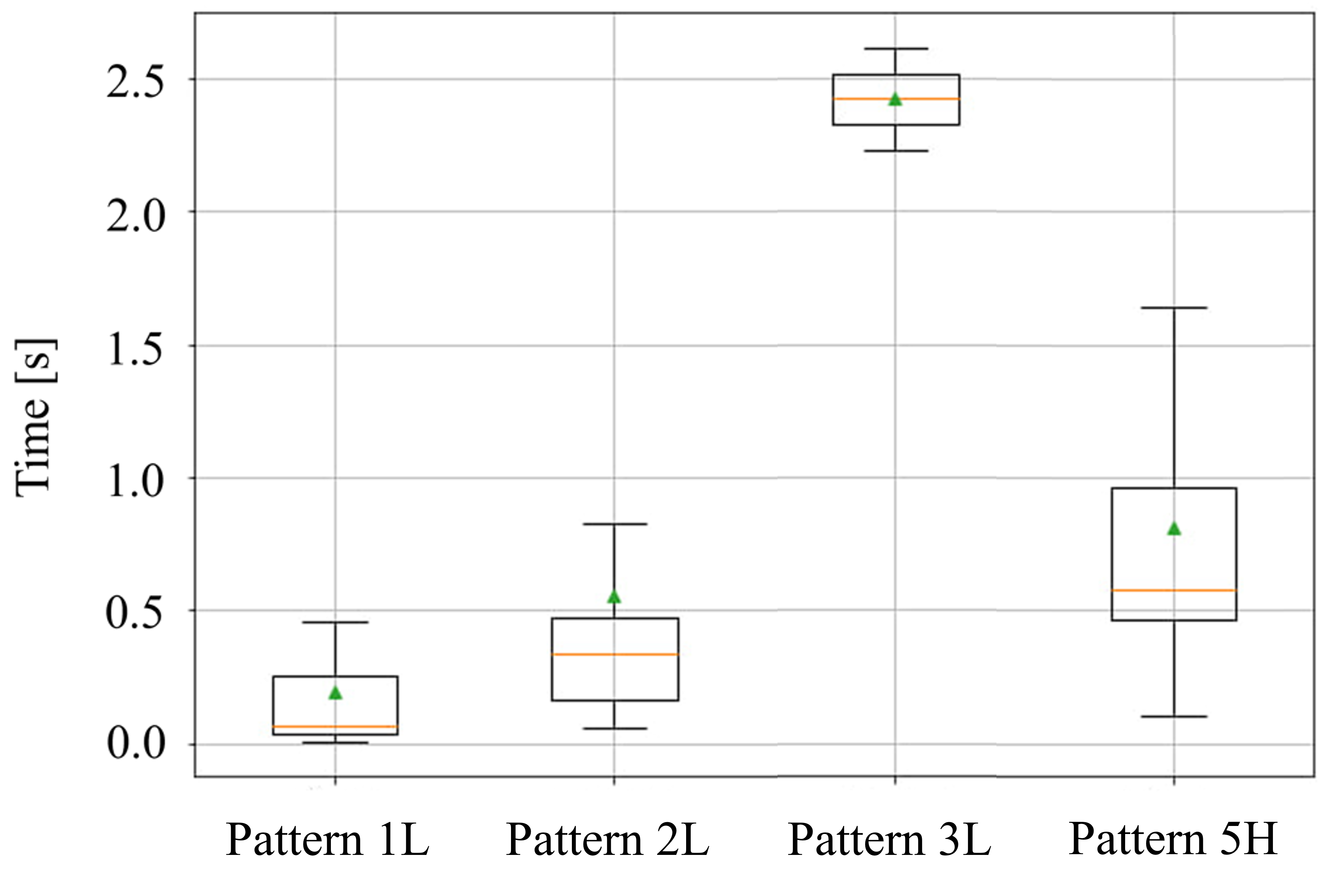}
  \qquad
  \caption{The response time for participants to remove their hands for the four patterns during the collaborative task.}
  \label{fig:time_response}
  \vspace*{-20pt}
\end{figure}

We measured the response time that the participants needed to get their hands out of the dangerous area. The time was measured between the moment the MoveTouch gives the haptic pattern and the moment users start to move their hands. The measured time for all users for each pattern can be seen in Fig. \ref{fig:time_response}. We can see that the mean response times were 0.24 seconds, 0.61 seconds, 0.85 seconds, and 2.41 seconds for the first, second, fourth, and third patterns, respectively. We can see that the response time of the fourth pattern is the highest among all the patterns.

\subsubsection{Conclusion}
The critical distance between the eye-covered participant and the robot's TCP was preserved throughout the collaboration, indicating the effectiveness of the designed vibrotactile patterns in guiding user actions when rendered on the wrist. The patterns remained discernible even when the participant's attention was focused on performing another task. The measured perception time of each pattern can serve as a design parameter to establish an upper limit for the robot's speed during collaboration, to assure that the robot does not critically approach the user, and thus, enhance user safety during collaborative tasks.

\section{Conclusion and Future Work}

This research presents an innovative wearable robotic system featuring both motion capture and a haptic guidance system designed for human-robot interaction. The system includes a wearable 2-DoF robot responsible for adjusting a marker held at its end-effector, ensuring its constant visibility to the camera. The haptic guidance system incorporates five vibration motors fixed on the wearable robot wrist rubber band to render vibrotactile patterns at the user's wrist to guide their hand movement. 

A total of ten haptic patterns were selected for experimentation on both the volar and dorsal parts of the wrist. A user study was conducted to evaluate the recognition and discriminability of these patterns. 
The average recognition rate of all patterns on the dorsal (up) part of the wrist was 66.6\%, while on the volar (down) side of the wrist, it was 75.64\%. In addition, almost all the vibration patterns were better recognized on the volar part of the wrist. Consequently, the decision was made to fix the vibration motors at the volar part of the wrist. Four patterns with high recognition rates were selected for further study.

Another experiment was made to evaluate the integration of the system into a real-world collaborative task incorporating the four selected patterns. We report that the critical distance had not been violated during the whole experiment and that the users were efficiently responding to the vibrotactile patterns. We also measured the perception time of each pattern and it was in the range 0.24-2.41 seconds.

For future work, we intend to expand the testing of the Movetouch robot across a broader user base and in diverse tasks to better evaluate the effectiveness of the proposed guidance system. We also aim to study how to enhance the recognition rate of the patterns by studying the effect of the pattern frequency. Additionally, we seek to minimize user response times to suggested patterns, considering that some patterns are perceived much faster than others.

\section*{Acknowledgements} 
The research reported in this publication was financially supported by the Russian Science Foundation grant No. 24-41-02039.
%
%
%


\end{document}